\documentclass{article}

\usepackage{arxiv}

\usepackage[utf8]{inputenc} 
\usepackage[T1]{fontenc}    
\usepackage{hyperref}       
\usepackage{url}            
\usepackage{booktabs}       
\usepackage{amsfonts}       
\usepackage{nicefrac}       
\usepackage{microtype}      
\usepackage{lipsum}
\usepackage{xcolor}
\usepackage{graphicx}

\usepackage{multirow}
\usepackage{graphicx}
\usepackage{amsmath}

\title{Texture Based Classification of High Resolution Remotely Sensed Imagery using Weber Local Descriptor}

\author{
  Decky Aspandi Latif\thanks{corresponding author, decky.aspandi.latif@gmail.com} \\
  Department of Computer Engineering \\
  King Mongkut's University of Technology Thailand\\
   \And
  Sally Goldin \\
  Department of Computer Engineering\\
  King Mongkut's University of Technology Thailand\\
  \And 
  Preesan Rakwatin\\
  Research and Development Group\\
  Geo-informatics and Space Technology Development Agency (GISTDA)\\
  \And 
  Kurt Rudahl \\
  Department of Computer Engineering\\
  KMUTT Geospatial Engineering and InnOvation Centre (kGeo), Geoinformation\\
  King Mongkut's University of Technology Thailand\\
}

\begin{document}
\maketitle

\begin{abstract}
Traditional image classification techniques often produce unsatisfactory results when applied to high spatial resolution data because classes in high resolution images are not spectrally homogeneous. Texture offers an alternative source of information for classifying these images. This paper evaluates a recently developed, computationally simple texture metric called Weber Local Descriptor (WLD) for use in classifying high resolution QuickBird panchromatic data. We compared WLD with state-of-the art texture descriptors (TD) including Local Binary Pattern (LBP) and its rotation-invariant version LBPRIU. We also investigated whether incorporating VAR, a TD that captures brightness variation, would improve the accuracy of LBPRIU and WLD. We found that WLD generally produces more accurate classification results than the other TD we examined, and is also more robust to varying parameters. We have implemented an optimised algorithm for calculating WLD which makes the technique practical in terms of computation time. Overall, our results indicate that WLD is a promising approach for classifying high resolution remote sensing data.
\end{abstract}


\section{Introduction}
As high resolution satellite imagery becomes more common, remote sensing applications can benefit from previously unavailable levels of detail. However, many traditional digital image analysis techniques produce unsatisfactory results when applied to high spatial resolution data~\cite{kim2008object}. In particular, classification algorithms such as maximum likelihood which are based on per-pixel spectral information often perform poorly ~\cite{carleer2005assessment, myint2011per}. At high spatial resolution, areas belonging to a particular class do not necessarily have homogeneous spectral characteristics. That is, adjacent pixels frequently have very different colour or brightness.

Texture provides an alternative information source for classifying high resolution imagery. High resolution images contain rich geometric structure with repeating variations in brightness, which may be regular within regions~\cite{zhang2011object}. Texture based classification incorporates information about the distribution of brightness or colour in the neighbourhood of the pixel of interest into the classification decision process.
Many approaches for classifying images based on texture have been investigated,   including Grey Level Co-occurrence Matrix (GLCM)\cite{haralick1973textural}, Wavelet-based Transform~\cite{ranchin1993wavelet,pohl1998review}, Gabor Filter~\cite{chang1993texture,risojevic2011gabor,clausi2000designing} and Scale Invariant Feature Transform (SIFT)~\cite{lowe2004distinctive,xu2012scale}. These techniques have shown promise but are rarely practical in operational contexts because all are very computationally intensive. 

Recently, several less computationally complex texture descriptors (TD) have been proposed, all of which compute a measure of variation in the immediate neighbourhood of the focus pixel, then construct a histogram of these measures within a somewhat larger window. The Local Binary Pattern (LBP) texture descriptor introduced by~\cite{ojala1996comparative} describes the local characteristics of the texture by comparing the centre pixel with its neighbours, then coding each comparison as either one (centre pixel is brighter than its neighbour) or zero (centre pixel is darker). This produces a small, fixed number of possible patterns of zeros and ones; the histogram of these patterns within a larger neighbourhood becomes the texture value for the focus pixel. This method has gained considerable support and has been successfully applied in texture analysis~\cite{ojala2001texture}, in face recognition~\cite{ahonen2006face}, and in image segmentation~\cite{li2008optimum}. However, LBP is limited due to the small size of the moving window. It is also not rotation-invariant and it cannot discriminate between textures defined by varying contrast. The first two problems can be addressed by combining several neighbourhood sizes to create a multi-resolution descriptor, and by adding rotation invariance to create the LBPRIU descriptor~\cite{ojala2002multiresolution}. Musci et al.~\cite{musci2013assessment} created a contrast sensitive LBPRIU by concatenating its histogram with the histogram from the Variance Texture Descriptor (VAR), creating the LBPRIU\_VAR descriptor. Although these efforts have produced encouraging results, the enhancements add to the computational costs, and thus reduce the simplicity that makes LBP so attractive.

A new texture descriptor called Weber Local Descriptor (WLD) was presented by in \cite{chen2009wld}. This TD uses ratios rather than simple comparisons and captures two aspects of texture, gradient orientation and central pixel value differentiation. WLD has been shown to rival some other classification methods based on texture such as SIFT and Gabor Filter~\cite{chen2009wld}. To the best of our knowledge, few remote sensing studies have utilised WLD and none has focused on texture classification in high resolution imagery. In this research, we apply the WLD method to remotely sensed images, employing a supervised pixel-wise classification approach based on a minimum distance metric. We compare the classification accuracy of WLD with LBP and several variants and examine the effects of neighbourhood size. As baselines, we also perform classifications on our test data using a classical GLCM-based texture approach and a traditional, non-texture-based maximum likelihood classification (MLC). Additional experimental conditions test the effectiveness of the Variance TD texture descriptor (VAR), both independently and combined with WLD, in images with significant illumination change within classes. Finally, we create optimised versions of the LBP and WLD algorithms which compute classified results sufficiently quickly to make these techniques practical for operational remote sensing applications.






\section{Methodology}
\label{sec:method}

\subsection{Image Data}
In this study, we used a panchromatic QuickBird image with 0.6 m spatial resolution acquired on 3 January 2007 and provided by the Geo-Informatics and Space Technology Development Agency (GISTDA). The image covers part of Pathumthani province in Thailand located at (14.083646 N,100.650593 E), WGS 1984 datum. Two 1024 x1024 pixel regions were extracted as test areas. Both regions include crop areas, naturally vegetated areas, barren areas and water bodies. Region 1 includes residential areas which represent artificial textures. Region 2 has texture areas affected by illumination changes that can be utilised to evaluate the potential benefits of introducing the VAR TD. We deliberately selected test areas with different characteristics to explore how this would affect accuracy. The test areas are shown in Figure \ref{fig:fig1}.

\begin{figure*}[t!]
  \centering
  \includegraphics[width=\linewidth]{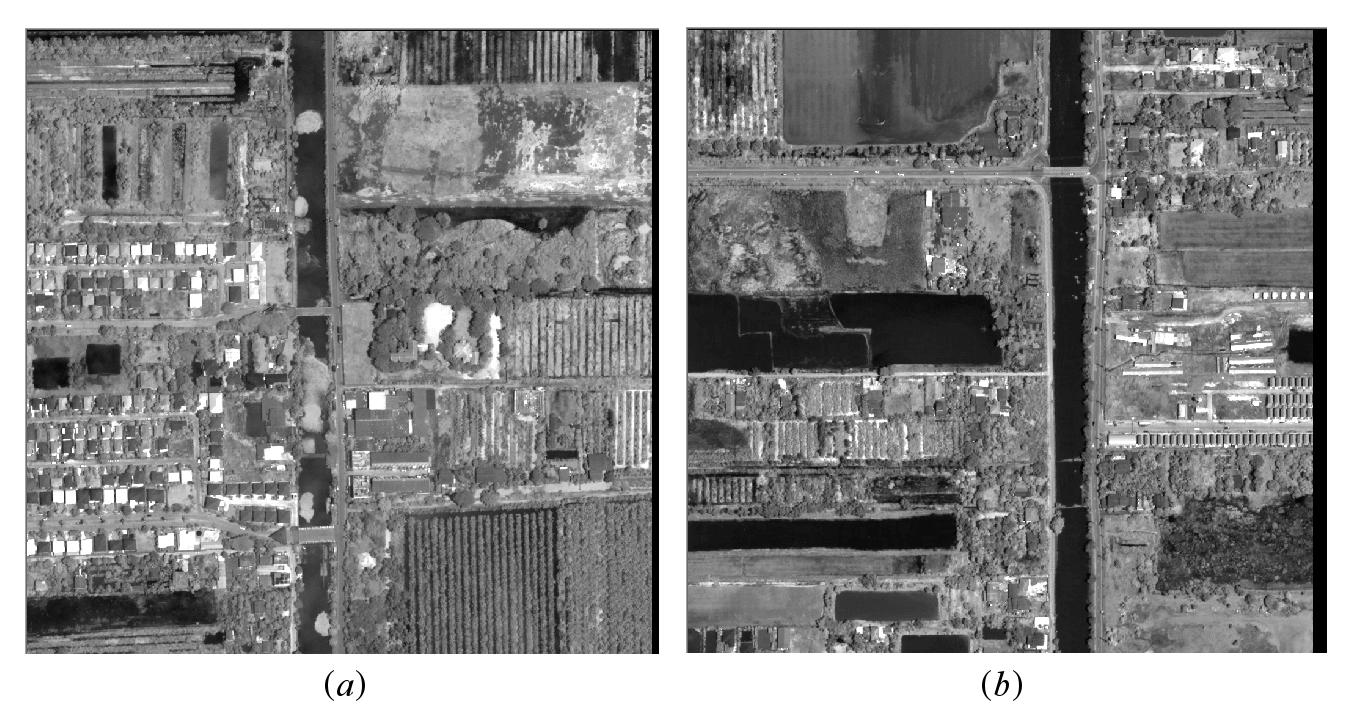}
  \caption{Study areas, (a) Region 1 and (b) Region 2.}
  \label{fig:fig1}
\end{figure*}

\subsection{Reference Data}
Given the time lapse since image acquisition, we did not expect ground surveys to yield accurate reference data. Instead, in order to create reference data for evaluation of classification accuracy, we asked an independent expert analyst with three decades of experience to locate and label major classes in each region using visual interpretation of both panchromatic and multispectral (2.4 m per pixel) QuickBird images. The expert identified ten classes based on various aspects of the image appearance, including colour, texture and spatial context, by outlining polygons surrounding each area representing a class. In total, there were 44 and 47 reference polygons identified for Region 1 and Region 2 respectively.

Some of the expert’s ten classes exhibited very inconsistent textures while others were too small to provide both training and test pixels. For our experiments, we selected four texture classes per region for a total of five texture classes.   Table \ref{tab:tab1} shows the selected texture classes for each region, with their pixel counts and other statistics.

\begin{table}[h!]
\begin{center}
\begin{tabular}{|l|c|c|c|c|c|c|c|}
\hline
\multirow{2}{*}{No.} & \multirow{2}{*}{Texture Class} & \multicolumn{3}{c|}{Region 1} & \multicolumn{3}{c|}{Region 2} \\ \cline{3-8} 
 &  & Pixel Count & Mean & SD & Pixel Count & Mean & SD \\ \hline
1. & Residential & 115142 & 49.33 & 14.81 & 0 & - & - \\ \hline
2. & Orchard & 176459 & 39.75 & 9.24 & 74720 & 39.55 & 10.35 \\ \hline
3. & Water & 67534 & 27.42 & 9.28 & 147822 & 17.73 & 4.71 \\ \hline
4. & Barren & 100061 & 47.63 & 8.59 & 121952 & 39.48 & 7.64 \\ \hline
5. & Crop & 0 & - & - & 145214 & 33.96 & 5.78 \\ \hline
\end{tabular}%
\caption{Statistics of Texture Classes.}
\end{center}
\label{tab:tab1}
\end{table}

\subsection{Training Samples}
We extracted training samples for each class by defining square areas within texturally homogeneous regions of the study images. Within each training area we calculated the pixel-wise TD for each pixel, then assembled the histogram of TD values as the standard model for that class. We considered three sample sizes: 30 x 30, 40 x 40 and 50 x 50.

To determine the best training area size, we conducted a preliminary classifications using each size using the state-of-the-art LBPRIU TD~\cite{ojala2002multiresolution} and selected the sample size with the highest accuracy. Training areas of 50 x 50 (2500 pixels) provided the best accuracy for region 1; 30 x 30 (900 pixel) training areas were best for region 2. The difference reflects the fact that texture units in region 1 tend to be larger than those in region 2. We chose four training areas of the appropriate size for each class, resulting in 10 000 training pixels per class for Region 1 and 3 600 training pixels per class for Region 2. TD values for all training pixels for a texture class were combined into a single histogram.

\subsection{Texture Descriptors} 

All the primary TD features tested in this study have a similar structure. They require two main parameters: the radius of the neighbourhood around the focus pixel (R) and the window size for aggregating individual pixel TD values into a histogram (termed 'ring size' in the literature). An R value of 1 means that the neighbourhood will involve the pixel's eight immediate neighbours, hence this situation is referred to as 'P:8,R:1'. A ring size of 40 means that each histogram will contain 40 x 40 or 1600 pixels.

The TD value for a particular pixel is calculated by comparing its brightness value with its neighbours to calculate one or two (in the case of WLD) numerical values. The texture within the window is characterised by creating a histogram that records the frequencies of each TD value, for all pixels in the window. The different TD have different methods for calculating the value(s) for an individual pixel.

\subsubsection{Local Binary Pattern (LBP)}
The LBP value for the centre pixel in a neighbourhood is calculated using the following equation~\cite{ojala1996comparative}:
\begin{equation}
    LBP_{P,R} = \sum_{i=0}^{p-1} u(t_i - t_c) x 2^i
\end{equation}
where P is the total number of neighbouring pixels and R is the radius used to form the circularly symmetric set of neighbours. When we are examining a pixel's immediate neighbours, P = 8, R = 1, ti is the brightness of a neighbour pixel while tc is the centre pixel brightness. The function $u(t_i - t_c )$ is a step function, where $u(x) = 1$ when $x \geq  0 ; else, u( x) = 0$. The formula produces an integer between 0 and 2P-1. If the values of R and P increase, changing the effective spatial resolution of the descriptor, the maximum possible value of the TD and hence the number of bins in the window histogram also increases. The power of 2 term in equation (1) means this increase is exponential. 

\subsubsection{Local Binary Pattern Rotation Invariant (LBPRIU)}
The binary pattern produced by LBP depends on the pixel values in each spatial neighbourhood. When the image is rotated, the pixel neighbourhood is also rotated. This will cause the binary pattern to move along the perimeter of the circle around the centre pixel. Hence, the LBP histogram for a texture will change if that texture is rotated, which may not be desirable.

LBPRIU~\cite{ojala2002multiresolution} eliminates this rotation effect by grouping LBP values according to the pattern of transitions between 0 and 1 in the binary representation of the value. For example, 000001002 and 000000102 will be mapped to the same LBPRIU group because there are two transitions, from 0 to 1 and 1 to 0, for both of these patterns. Only the position of the transitions differs, a variation that would result from rotating a texture. The mapping procedure uses the equation below: 

\begin{equation}
    LBP_{P,R}^{riu2} = \binom{\sum_{p=0}^{p-1}s(t_p-t_c) \; if \; U(LBP_{P,r}) \leq  2)}{P+1 \; otherwise,} 
\end{equation}

P and R represent the number of pixels and radius of the neighbourhood as before. The function s is the signum fuction, $s(x) = 1 if x >=0, else \; s(x) = -1$. The uniformness function U is defined by equation (3).
\begin{equation}
U(LBP_{P,R}) = (s(t_{p-1}-t_{c})-s(t_0-t)c))+sum_{p=1}^{p-1}(s(t_p-t_c)-s(t_{p-1}-t_c))
\end{equation}

The value generated by U depends on the number and pattern of binary digit transitions. The mapping procedure implemented by LBPRIU collects LBP values with U <= 2 into P distinct classes. LBP values with U > 2 are grouped into one miscellaneous class labelled as (P+1). The histogram for the window reflects the counts of pixels in each class. The number of bins in the LPBRIU histogram increases with the neighbourhood size, but not exponentially as in LBP.

\subsubsection{VAR and LBPRIU\_VAR (Variance)}
Both the LBP and the LBPRIU descriptors are invariant to monotonic grey scale changes and consequently do not capture either absolute brightness or contrast information. Ojala et al. \cite{ojala2002multiresolution} proposed a local contrast descriptor, denoted as $VAR_{P,R}$, which is also rotation invariant, defined as:

\begin{center}

\begin{equation}
    VAR_{P,R}(w) = \frac{1}{P} \sum_{p=0}^{p-1}(t_p-u)^2 
\end{equation}

where, 
\begin{equation}
    u = \frac{1}{P} \sum_{p=0}^{p-1}t_p
\end{equation}

\end{center}

$VAR_{P,R}$ is an approximation of local variance, which can be computed efficiently if performed concurrently with the computation of $LBPRIU_{P,R}$ . LBPRIU and VAR can be combined to produce the LBPRIU\_VAR texture descriptor, an extension of LBPRIU that also captures the contrast feature of texture. This can be achieved by concatenating the VAR histogram with the LBPRIU histogram~\cite{musci2013assessment}.

\subsubsection{Weber Local Descriptor (WLD}
Weber Local Descriptor (WLD) was introduced b~\cite{chen2009wld}. This descriptor is inspired by Weber's Law in physiological psychology, which states that the minimum distinguishable change of a perceptual stimulus is a constant fraction of the original stimulus. When the change is smaller than this constant, a human being would consider the stimulus as background noise rather than a valid signal. WLD adopts this idea by using a ratio to represent differences in the values of neighbouring pixels.

WLD consists of two components: differential excitation $(\xi)$ and orientation $(\theta)$. $\xi$ is a function of the Weber fraction (i.e., the intensity differences of a pixel's neighbours relative to its own intensity). $\theta$ is a gradient orientation of the current pixel. The excitation is defined by: 

\begin{equation}
\xi (x_c) = arctan \begin{bmatrix}
\sum_{i=0}^{n-1} \frac{I_i-I_c}{I_c}
\end{bmatrix}
\end{equation}

where $n$ is the number of neighbouring pixels and $I_i$ is pixel value at position i. Pixels are numbered clockwise starting with 0 at the upper left. The gradient orientation is defined by: 

\begin{equation}
\theta (x_c) = arctan \begin{bmatrix}
\sum_{i=0}^{n-1} \frac{I_7-I_3}{I_5-I_1}
\end{bmatrix}
\end{equation}

Based on these two terms, a two dimensional joint histogram can be constructed, followed by conversion to a one dimensional histogram, which is the WLD descriptor for a window. For details of the conversion process see \cite{chen2009wld} and \cite{noauthororeditor2014texture}. 

\subsubsection{Texture based pixel-wise supervised classification}
We used texture based pixel-wise supervised classification to classify each pixel into a particular group by considering the texture in its neighbourhood. The histograms of the training areas described in section 2.3 constitute the a priori definitions of our classes. Each pixel in the image is classified by calculating the histogram of TD values within a window centred on that pixel, calculating a similarity measure against the training histogram for each class and placing the pixel into the most similar class~\cite{topi2000texture}. Based on the literature plus initial experiments, we chose the Bhatacharya distance as our similarity metric. Bhatacharya distance measures the distance or difference between two histograms. In remote sensing this metric is often used to assess spectral separability of classes~\cite{richards1999remote}, but it can also measure similarity between the TD histogram calculated for a pixel and the TD histogram for a training sample~\cite{tuzel2006region}. Bhatacharya distance is calculated as follows: 

\begin{equation}
BD(H^1, H^2) = -ln \left ( \sum_{x=x} \sqrt{H^{1,i}\; \times \; H^{2,i}} \right ) 
\end{equation}

$H^{1,i}$ and $H^{2,i}$ are the counts for bin i in the first and second histogram respectively. This distance metric has moderate computational complexity and produces bounded values between 0 to 1, where smaller values imply greater similarity between two histograms. 

\section{Experiments}

\subsection{Experimental Settings}
Our main experiments varied the following conditions:
\begin{enumerate}
    \item Texture descriptor: LPBP, LBPRIU, WLD, VAR, LBPRIU\_VAR and WLD\_VAR. The first three TD do not capture contrast differences between texture classes. The second three attempt to involve contrast features by incorporating variance. WLD\_VAR is a new TD first investigated by Latif~\cite{noauthororeditor2014texture} that combines WLD and VAR into a single histogram, analogous to LBPRIU\_VAR.
    \item Scale: We calculated each TD at several scales (P:8,R:1; P:16,R:2; P:24,R:3) to examine whether larger scales might generate higher classification accuracy as has been reported by some authors. We also examined multi-resolution classification, by concatenating the histograms of all scales in a single classification. 
\end{enumerate}
We used confusion matrices to calculate the overall accuracy and Kappa coefficient~\cite{cohen1960coefficient} for each classification condition. After removing pixels used for training, all pixels labelled by the expert image analyst in each of our selected classes were used as test pixels. Finally, we also recorded the elapsed time for each experimental condition. 

\subsection{Tools}
\label{sec:tools}
The texture-extraction and classification software used in this study was written by the first author in GNU C++ using the Dragon Programmer's Toolkit™. Experiments were executed on an Intel Centrino dual core processor 2.1 Ghz, with 4 GB of RAM, running Ubuntu Linux 12.04. Dragon/ips®~\cite{rudahl2017opendragon} was used to perform standard remote sensing procedures, such as defining test areas, clipping, calculation of accuracy, image display, etc. \

\subsection{Experimental Results}

\subsubsection{LBP and LBPRIU versus WLD: general accuracy comparison.} 
Tables 2 and 3 present the overall classification accuracy using LBP, LBPRIU, and WLD texture descriptors on regions 1 and 2 respectively, for various scales. The Kappa coefficients show a similar pattern, although their values as expected are lower than overall accuracy.

\begin{table}[h!]
\begin{center}
\begin{tabular}{|c|c|c|c|}
\hline
P,R & LBP (\%) & LBPRIU (\%) & WLD (\%) \\ \hline
8,1 & 73.94 & 62 & \textbf{84.55} \\ \hline
16,2 & \textbf{83.59} & 65.01 & 82.05 \\ \hline
24,3 & 70.48 & 60.1 & \textbf{73.29} \\ \hline
Multi-scale & \textbf{84.49} & 66.52 & 81.76 \\ \hline
Mean & 78.13 & 63.41 & \textbf{80.41} \\ \hline
\end{tabular}%
\caption{Overall percent correct classification from LBP, LBPRIU and WLD on region 1.}
\end{center}
\end{table}

\begin{table}[h!]
\begin{center}
\begin{tabular}{|c|c|c|c|}
\hline
P,R & LBP (\%) & LBPRIU (\%) & WLD (\%) \\ \hline
8,1 & 66.91 & 64.33 & \textbf{67.63} \\ \hline
16,2 & 52.99 & 60.84 & \textbf{68.49} \\ \hline
24,3 & 35. 99 & 51.01 & \textbf{62.73} \\ \hline
Multi-scale & 62.13 & 60.03 & \textbf{71.56} \\ \hline
Mean & 54.51 & 59.05 & \textbf{67.60} \\ \hline
\end{tabular}%
\caption{Overall percent correct classification from LBP, LBPRIU and WLD on region 2.}
\end{center}
\end{table}

These tables show that WLD consistently outperforms both generic LBP and LBPRIU with the exception that in region 1 basic LBP with the spatial resolution of 2 and multi-scale are slightly more accurate than WLD. In these resolutions, however, LBP is extremely slow. 

Note that the accuracies of all TD are better for region 1 than region 2. We believe this is due to the fact that the illumination in the second region is less consistent. This makes the classification problem harder than in the first region where textures are relatively homogeneous in terms of brightness. Another notable result is that adding rotation invariance to LBP to create LBPRIU does not provide any consistent improvements on accuracy, especially in region 1, where basic LBP produces better accuracy than LBPRIU overall. In region 2, LBPRIU sometimes produced more accurate results than LBP. This may be due to the fact that most of the textured areas have similar orientations. In region 2, LBPRIU sometimes produced more accurate results than LBP, but was not consistently superior.	

\subsubsection{LBP and LBPRIU versus WLD: multi-resolution analysis}
Tables 2 and 3 also allow us to examine the accuracy of LBP, LBPRIU and WLD when applied at different scales. Although accuracy does vary with scale, there are few clearfew consistent patterns within a TD. Generally, the spatial resolution of 1 for each TD produces more accurate results than other scales. Moreover, in contrast to the reported results in \cite{chen2009wld} we found no improvement with the multi-scale version of WLD compared to the single scale version. Instead the multi-scale results are close to the mean results across scales for each TD.

We can make other observations. First, WLD provides more consistent results across different resolutions. In other words, WLD is less sensitive to variation of parameters. This robustness is a desirable feature. Second, the accuracy of different resolutions very likely depends on the characteristics of textures in the specific images. Thus it is possible that scales greater than 1 would provide improved accuracy for images with larger texture units.

\subsection{The effect of variance on WLD and LBPRIU: general accuracy comparison.}
Tables 4 and 5 compare the overall accuracy for the single TD metrics LBPRIU, WLD and VAR, plus the compound metrics LBPRIU\_VAR and WLD\_VAR. The VAR TD by itself is capable of classifying images quite accurately, with a mean of about 67\% for both regions. However in region 1, at least, WLD continues to provide higher accuracy in most cases. Furthermore, combining LBPRIU with VAR noticeably increases accuracy compared to plain LBPRIU, by more than 10\% in some cases. This result agrees with~\cite{musci2013assessment}. However, in general VAR makes no contribution to WLD, at least in region 1. In region 2, where the image data exhibit illumination changes within texture regions, VAR slightly enhances the overall WLD results in all scale, with the maximum increase in accuracy of about 7

\begin{table}[h!]
\begin{center}
\begin{tabular}{|c|c|c|c|c|c|}
\hline
P,R & LBPRIU (\%) & WLD(\%) & VAR(\%) & LBPRIUVAR(\%) & WLDVAR(\%) \\ \hline
8,1 & 62 & \textbf{84.55} & 69.68 & 69.35 & 79.13 \\ \hline
16,2 & 65.01 & \textbf{82.05} & 66.87 & 69.42 & 80.53 \\ \hline
24,3 & 60.1 & 73.29 & 66.63 & 71.22 & \textbf{79.85} \\ \hline
Multi-scale & 66.52 & \textbf{81.76} & 68.21 & 70,36 & 80.59 \\ \hline
Mean & 63.41 & \textbf{80.41} & 67.85 & 70.09 & 80.04 \\ \hline
\end{tabular}%
\caption{ Percent correct classification of region 1 using several TD combined with VAR.}
\end{center}
\end{table}

\begin{table}[h!]
\begin{center}
\begin{tabular}{|c|c|c|c|c|c|}
\hline
P,R & LBPRIU(\%) & WLD(\%) & VAR(\%) & LBPRIUVAR(\%) & WLDVAR(\%) \\ \hline
8,1 & 64.33 & 67.63 & 68.25 & 68.37 & \textbf{75.34} \\ \hline
16,2 & 60.84 & 68.49 & 66.23 & 66.41 & \textbf{70.69} \\ \hline
24,3 & 51.01 & 62.73 & 62.59 & 62.92 & \textbf{65.18} \\ \hline
Multi-scale & 60.03 & 71.56 & 66.05 & 66.17 & \textbf{71.60} \\ \hline
Mean & 59.05 & 67.60 & 65.78 & 65.97 & \textbf{70.70} \\ \hline
\end{tabular}
\caption{Percent correct classification of region 2 using several TD combined with VAR.}
\end{center}
\end{table}

\subsubsection{The effect of variance on WLD and LBPRIU: multi-resolution analysis.} We can also consider the accuracy of WLD, LBPRIU, VAR, LBPRIU\_VAR and WLD\_VAR when applied at different scales. Tables 4 and 5 show patterns similar to the previous multi-resolution analysis. Using different spatial scales affects accuracy, but not in any consistent way. The multi-scale version is not more accurate than the single scale classifications, but generates the results close to the mean of all resolutions.

\subsection{Comparison with GLCM and maximum likelihood classification}
To further examine the performance of WLD and the other histogram-based TDs, we classified our test data using two additional methods. Specifically, we conducted a texture-based classification using the classical GLCM approach. We also performed a traditional, non-texture-oriented Maximum Likelihood classification. 

Software for the GLCM classification was implemented by the first author using the tools described in Section~\ref{sec:tools}.   We selected GLCM parameters values based on the most commonly reported values in the remote sensing literature \cite{clausi2002rapid,soh1999texture,petrou2021image,clausi2002rapid}. We used four orientation angles (0, 45, 90, 135), seven GLCM statistics ( Entropy, Energy, Homogeneity, Dissimilarity, Variance, Shade , Correlation and Contrast), a window size of 7x7, GLCM quantization levels of 32x32 and 64x64, and distance values 1, 2 and 3. 
MLC was performed using Dragon/ips®. Since this method requires multiple bands, we used the four multispectral bands of QuickBird data that correspond to the panchromatic image in our primary experiment. Training signature statistics (mean, variance and covariance) were calculated from the same training pixels used in experiments on the primary TDs.

Table 6 shows the results of these traditional classification approaches, compared to the average results from the primary TDs, for regions 1 and 2. The accuracies for both GLCM and MLC Maximum Likelihood are quite high in Region 1 but lower in Region 2. This is consistent with the results from earlier experiments. However, as expected, the traditional methods produce less accurate results than the primary TDs of interest in every case, with a maximum difference of about 10\% for MLC and 13\% for GLCM compared with the best accuracy results of each TD. Full details of all results reported in this article (corresponding confusion matrix, omission errors, etc.) can be found at https://sites.google.com/a/mail.kmutt.ac.th/upload/.

It is important to note that the spatial resolution of the data used in the maximum likelihood classification (2.4 m multispectral Quickbird) is lower than that used for the texture-based classifications (0.6 m panchromatic Quickbird). Each pixel in the multi-spectral bands corresponds to nine pixels in the panchromatic image. This should favour maximum likelihood, since the influence of within-class spectral diversity will be reduced to some extent compared to the higher resolution. Nevertheless, our results suggest that we might improve overall accuracy by incorporating spectral information into our texture descriptors.

\begin{table}[h!]
\begin{center}
\begin{tabular}{|c|c|c|}
\hline
Overall Accuracy & Region 1 \% correct (Kappa) & Region 2 \% correct (Kappa) \\ \hline
GLCM & 71.53 (0.60) & 61 (0.48) \\ \hline
Maximum Likelihood & 74.48 (0.65) & 62.29 (0.49) \\ \hline
LBP & 78.13 (0.71) & 54.51 (0.37) \\ \hline
LBPRIU & 63.41 (0.50) & 59.05 (0.46) \\ \hline
WLD & \textbf{80.41 (0.73)} & 67.60 (0.57) \\ \hline
VAR & 67.85 (0.56) & 65.78 (0.55) \\ \hline
LBPRIU\_VAR & 70.09 (0.59) & 65.97 (0.55) \\ \hline
WLD\_VAR & 80.04 				(0.72) & \textbf{70.70 (0.61)} \\ \hline
\end{tabular}
\caption{ Result for GLCM and maximum likelihood in region 1 and region 2 compared with average result of earlier TDs classification.}
\end{center}
\end{table}

\subsection{Computation time}

\begin{figure*}[t!]
  \centering
  \includegraphics[width=\linewidth]{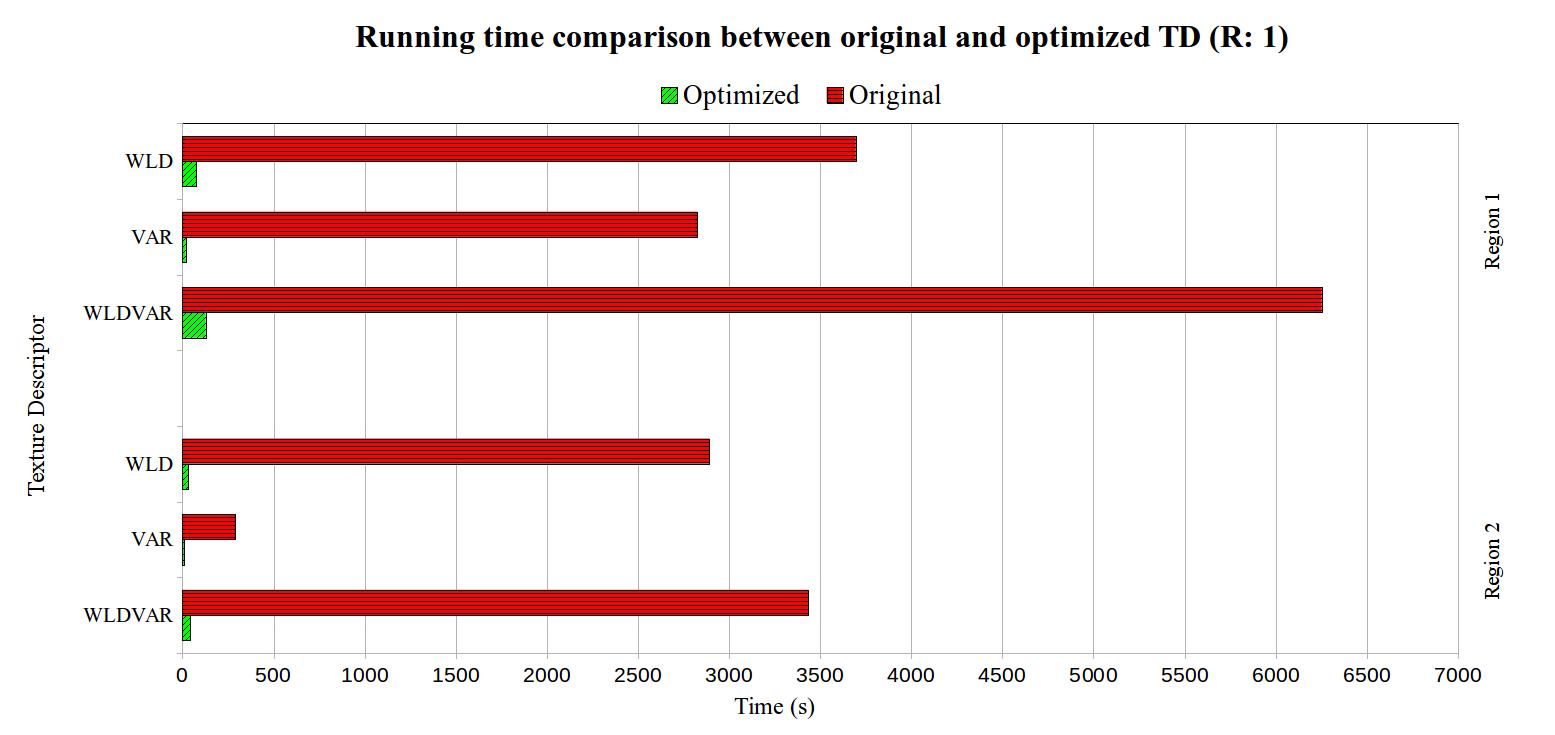}
  \caption{Comparison of Pixelwise classification running time between original and optimised texture descriptors.}
  \label{fig:fig1}
\end{figure*}

Our initial implementation of TD algorithms, reported in Latif~\cite{noauthororeditor2014texture}, was a naive translation from equations logical structure to code. We found that the computation time for the different TD varied widely. In particular, LBP was the fastest TD at scale 1, but became impractically slow at higher resolutions. We attributed this to the exponentially increasing number of bins in the LBP histogram at higher scales, all of which must be scanned during similarity computations. The computation time for LBPRIU did not grow in the same manner, partly because the LBPRIU remapping produces a small number of histogram bins regardless of the scale. WLD was somewhat slower than LBP or LBPRIU at scale 1, but provided more consistent run times across scales and regions.

Leaving out the multi-resolution case, the average run time for WLD was 4 004 s for region 1 and 2 865 s for region 2. If we extrapolate from our 1024 x 1024 subsets to a more typical image size of 6000 x 6000 pixels, the processing time for a texture based classification using WLD would be more than 100 000 seconds (about 30 hours), which is not really practical for an operational environment.

However, when we reviewed our code, we noted that our original implementation repeated many calculations.. We redesigned our software to cache various intermediate values. Our optimisations produced a dramatic speed-up, as shown in Figure 2. For WLD, for instance, the classification time dropped to about 52 s for a 1024 x 1024 region. This translates to about half an hour for a full scene. Figure 2 shows only the P:8,R1 condition, but results were comparable for other resolutions.

This calculation time is much faster than GLCM-based classification which required about 750 s and 2 400 s for region 1 and region 2 respectively. Note that our implementation of GLCM was optimised using   similar caching strategies similar to those employed for other algorithms. Hence, our optimised implementation of WLD provides both high accuracy and relatively good computational performance. Further performance gains can probably be achieved by exploiting the inherent parallelism in the pixel-wise computations. 

\section{Discussion}
To our knowledge, this research is the first to apply WLD for texture based classification of high resolution remotely sensed images. The results are encouraging. In almost every case, WLD produced more accurate results than LBP, the state-of-the-art LBPRIU or traditional GLCM-based methods. Concatenating the VAR TD with the WLD TD improved accuracy further in an image with obvious within-class illumination variation, but had no effect in an image with more homogeneous illumination. Unlike LBP, WLD values do vary somewhat depending on pixel brightness. The VAR metric, by comparison, is sensitive to the range of brightness values around a pixel but not their absolute levels. We believe that VAR improves WLD performance in Region 2 because it ignores illumination variability within classes while incorporating the additional feature of contrast.

Our work also suggests that varying the scale of these TDs will not necessarily improve classification accuracy, and that using multiple scales concurrently provides no benefit. This conclusion is tentative since scale effects probably depend strongly on the images being classified. 
After demonstrating the superior accuracy of WLD, we optimised our software to improve computational performance by almost two orders of magnitude. Thus our implementation of WLD is practical for operational remote sensing applications.

This research is of course preliminary. We have not attempted statistical analyses of significance on due to limitations on available training data. Comparable experiments should be performed using other sensors and geographic areas, with ground truth based on actual surveys, rather than independent visual interpretation.

We see many interesting directions for future research. WLD and other binary TD approaches are currently limited to a single band of data. We are working on schemes to incorporate spectral information into a WLD-like TD. We are also investigating alternative methods for incorporating variance information into WLD, since simple histogram concatenation has a negative impact on computational performance. Finally, we hope to explore the use of WLD for segmentation and unsupervised classification of high resolution remotely sensed images. 









\bibliographystyle{unsrt}  
\bibliography{references}  

\begin{thebibliography}{10}

\bibitem{kim2008object}
Minho Kim, Bo~Xu, Marguerite Madden, and ISPRS.
\newblock Object-based vegetation type mapping from an orthorectified
  multispectral ikonos image using ancillary information.
\newblock {\em Proceedings of the GEOBIA}, 2008.

\bibitem{carleer2005assessment}
AP~Carleer, Olivier Debeir, and El{\'e}onore Wolff.
\newblock Assessment of very high spatial resolution satellite image
  segmentations.
\newblock {\em Photogrammetric Engineering \& Remote Sensing},
  71(11):1285--1294, 2005.

\bibitem{myint2011per}
Soe~W Myint, Patricia Gober, Anthony Brazel, Susanne Grossman-Clarke, and Qihao
  Weng.
\newblock Per-pixel vs. object-based classification of urban land cover
  extraction using high spatial resolution imagery.
\newblock {\em Remote sensing of environment}, 115(5):1145--1161, 2011.

\bibitem{zhang2011object}
Yanmei Zhang, Jixian Zhang, Guoman Huang, and Zheng Zhao.
\newblock Object-oriented classification of polarimetric sar imagery based on
  texture features.
\newblock In {\em 2011 International Symposium on Image and Data Fusion}, pages
  1--4. IEEE, 2011.

\bibitem{haralick1973textural}
Robert~M Haralick, Karthikeyan Shanmugam, and Its'~Hak Dinstein.
\newblock Textural features for image classification.
\newblock {\em IEEE Transactions on systems, man, and cybernetics},
  (6):610--621, 1973.

\bibitem{ranchin1993wavelet}
Thierry Ranchin and Lucien Wald.
\newblock The wavelet transform for the analysis of remotely sensed images.
\newblock {\em International Journal of Remote Sensing}, 14(3):615--619, 1993.

\bibitem{pohl1998review}
Cle Pohl and John~L Van~Genderen.
\newblock Review article multisensor image fusion in remote sensing: concepts,
  methods and applications.
\newblock {\em International journal of remote sensing}, 19(5):823--854, 1998.

\bibitem{chang1993texture}
Tianhorng Chang and C-CJ Kuo.
\newblock Texture analysis and classification with tree-structured wavelet
  transform.
\newblock {\em IEEE Transactions on image processing}, 2(4):429--441, 1993.

\bibitem{risojevic2011gabor}
Vladimir Risojevi{\'c}, Snje{\v{z}}ana Momi{\'c}, and Zdenka Babi{\'c}.
\newblock Gabor descriptors for aerial image classification.
\newblock In {\em International Conference on Adaptive and Natural Computing
  Algorithms}, pages 51--60. Springer, 2011.

\bibitem{clausi2000designing}
David~A Clausi and M~Ed Jernigan.
\newblock Designing gabor filters for optimal texture separability.
\newblock {\em Pattern Recognition}, 33(11):1835--1849, 2000.

\bibitem{lowe2004distinctive}
David~G Lowe.
\newblock Distinctive image features from scale-invariant keypoints.
\newblock {\em International journal of computer vision}, 60(2):91--110, 2004.

\bibitem{xu2012scale}
Yong Xu, Sibin Huang, Hui Ji, and Cornelia Ferm{\"u}ller.
\newblock Scale-space texture description on sift-like textons.
\newblock {\em Computer Vision and Image Understanding}, 116(9):999--1013,
  2012.

\bibitem{ojala1996comparative}
Timo Ojala, Matti Pietik{\"a}inen, and David Harwood.
\newblock A comparative study of texture measures with classification based on
  featured distributions.
\newblock {\em Pattern recognition}, 29(1):51--59, 1996.

\bibitem{ojala2001texture}
Timo Ojala, Kimmo Valkealahti, Erkki Oja, and Matti Pietik{\"a}inen.
\newblock Texture discrimination with multidimensional distributions of signed
  gray-level differences.
\newblock {\em Pattern Recognition}, 34(3):727--739, 2001.

\bibitem{ahonen2006face}
Timo Ahonen, Abdenour Hadid, and Matti Pietikainen.
\newblock Face description with local binary patterns: Application to face
  recognition.
\newblock {\em IEEE transactions on pattern analysis and machine intelligence},
  28(12):2037--2041, 2006.

\bibitem{li2008optimum}
Ma~Li and Richard~C Staunton.
\newblock Optimum gabor filter design and local binary patterns for texture
  segmentation.
\newblock {\em Pattern Recognition Letters}, 29(5):664--672, 2008.

\bibitem{ojala2002multiresolution}
Timo Ojala, Matti Pietikainen, and Topi Maenpaa.
\newblock Multiresolution gray-scale and rotation invariant texture
  classification with local binary patterns.
\newblock {\em IEEE Transactions on pattern analysis and machine intelligence},
  24(7):971--987, 2002.

\bibitem{musci2013assessment}
Marcelo Musci, Raul~Queiroz Feitosa, Gilson~AOP Costa, and Maria
  Luiza~Fernandes Velloso.
\newblock Assessment of binary coding techniques for texture characterization
  in remote sensing imagery.
\newblock {\em IEEE Geoscience and Remote Sensing Letters}, 10(6):1607--1611,
  2013.

\bibitem{chen2009wld}
Jie Chen, Shiguang Shan, Chu He, Guoying Zhao, Matti Pietik{\"a}inen, Xilin
  Chen, and Wen Gao.
\newblock Wld: A robust local image descriptor.
\newblock {\em IEEE transactions on pattern analysis and machine intelligence},
  32(9):1705--1720, 2009.

\bibitem{noauthororeditor2014texture}
Decky Aspandi-Latif, Sally Goldin, and Sumaryono.
\newblock Texture based classification of high resolution remotely sensed
  imagery using weber local descriptor.
\newblock In {\em Asia GIS 2014 International Conference}, volume~10, 2014.

\bibitem{topi2000texture}
M{\"a}enp{\"a}{\"a} Topi, Pietik{\"a}inen Matti, and Ojala Timo.
\newblock Texture classification by multi-predicate local binary pattern
  operators.
\newblock In {\em Proceedings 15th International Conference on Pattern
  Recognition. ICPR-2000}, volume~3, pages 939--942. IEEE, 2000.

\bibitem{richards1999remote}
A~Richards~John and Jia Xiuping.
\newblock Remote sensing digital image analysis: an introduction, 1999.

\bibitem{tuzel2006region}
Oncel Tuzel, Fatih Porikli, and Peter Meer.
\newblock Region covariance: A fast descriptor for detection and
  classification.
\newblock In {\em European conference on computer vision}, pages 589--600.
  Springer, 2006.

\bibitem{cohen1960coefficient}
Jacob Cohen.
\newblock A coefficient of agreement for nominal scales.
\newblock {\em Educational and psychological measurement}, 20(1):37--46, 1960.

\bibitem{rudahl2017opendragon}
Kurt Rudahl and Sally~E Goldin.
\newblock Opendragon: software and a programmer’s toolkit for teaching remote
  sensing and geoinformatics.
\newblock {\em Open Geospatial Data, Software and Standards}, 2(1):1--7, 2017.

\bibitem{clausi2002rapid}
David~A Clausi and Yongping Zhao.
\newblock Rapid extraction of image texture by co-occurrence using a hybrid
  data structure.
\newblock {\em Computers \& Geosciences}, 28(6):763--774, 2002.

\bibitem{soh1999texture}
L-K Soh and Costas Tsatsoulis.
\newblock Texture analysis of sar sea ice imagery using gray level
  co-occurrence matrices.
\newblock {\em IEEE Transactions on geoscience and remote sensing},
  37(2):780--795, 1999.

\bibitem{petrou2021image}
Maria~MP Petrou and Sei-ichiro Kamata.
\newblock {\em Image processing: dealing with texture}.
\newblock John Wiley \& Sons, 2021.

\end{thebibliography}






\end{document}